%% file: main.tex
\documentclass[letterpaper]{article} 
\usepackage[preprint]{aaai2027}  
\usepackage[hyphens]{url}  
\usepackage{graphicx} 
\usepackage{natbib}  
\usepackage{caption} 
\usepackage{booktabs}
\usepackage{amsmath,amssymb}
\title{When Do Learned Diffusion Proposals Help Constraint Solving?\\
A Controlled Study on Continuous Algebraic Systems}
\author{
    Quang Bui\equalcontrib\textsuperscript{\rm 1,\rm 2},
    Sparsh Roy\equalcontrib\textsuperscript{\rm 1,\rm 3},
    Akash Gundimeda\equalcontrib\textsuperscript{\rm 1,\rm 4},
    Davin Yin\equalcontrib\textsuperscript{\rm 1,\rm 5}
}
\affiliations{
    \textsuperscript{\rm 1}SAID Lab\\
    \textsuperscript{\rm 2}American International School Vienna\\
    \textsuperscript{\rm 3}Hopewell Valley Central High School\\
    \textsuperscript{\rm 4}Centennial High School\\
    \textsuperscript{\rm 5}Harvard University\\
    quangbui@mit.edu, sparshr@mit.edu, akash\_g@mit.edu, davin\_yin@college.harvard.edu
}
\date{}

\begin{document}
\maketitle

\begin{abstract}
Solving a continuous algebraic constraint system requires two decisions: which values satisfy
the constraints, and which structural augmentation renders an unsolvable system solvable.
Classical solvers answer the first well and the second only by enumeration. On that discrete
decision, a candidate-conditioned repair ranker choosing among $K$ candidate augmentations
reaches the exhaustive-search ceiling at a fraction of the calls, outperforming random
(0.997 vs 0.236 balanced nonlinear menu accuracy; $p<10^{-70}$; $0.982\pm0.006$ across seeds)
and beating a budget-matched per-candidate probe on accuracy and cost. MARC turns such a
system into a factor graph, over which a graph-neural diffusion denoiser proposes assignments,
descent on an exact computer-algebra energy polishes them, and an exact symbolic checker
certifies solutions. Evaluations of diffusion-based proposal models rarely include one key
control: random multi-start under the same refinement budget. When applied to our system, this
control sharply curtails what the learned proposal contributes on the value decision. Does it
improve on random multi-start at choosing satisfying assignments? Only narrowly, in a
predictable regime. Across trapped low-dimensional families it ties with random restart, but
dominates in high dimension, where random search fails. Once variables couple, the advantage
is gone. Since all methods share the same polishing and checker, best-of-$K$ random
multi-start succeeds with probability exactly $1-(1-q(n))^K$, where $q(n)$ is single-start
reachability; a single measured constant, with no free parameters, reproduces the entire curve
(mean absolute error 0.012). The narrow favorable regime is not specific to our synthetic
families: across eight real-world systems in robotics, positioning, optimization, and algebra,
classical multi-start solved all eight, but none were in the learning-favorable regime. We delineate the regimes in which learned proposals improve solvers.
\end{abstract}

\section{Introduction}
\label{sec:intro}

Solving a system of algebraic constraints requires a pair of decisions, one continuous and
one discrete. The former asks which values satisfy the constraints; it is differentiable and
has been studied for six decades in numerical analysis. The latter asks which representation
renders those values tractable to find: which auxiliary variables, substitutions, or defining
relations should be introduced. Classical numerical solvers take the representation as given,
making the structural choice only through enumeration; on the value decision they are the
method to beat. This paper evaluates what a learned component is worth in each, under one
controlled protocol.

This paper reports one clear positive and the sequence of controls that produced it. The
positive is on the structural decision. A repair ranker that conditions on each
candidate-augmented graph, rather than on the instance alone, clears its matched controls on
certificate-grade nonlinear menus and reaches the exhaustive-enumeration ceiling at a small
fraction of the calls. It also keeps its value away from benchmarks of our own design: on hardened variants of named
real systems, a construction derived from the givens and selected on held-out failures repairs
failures that no budget-matched restart control reaches. The same controls cost us a claim on
the value decision, where learning pays only inside a regime we characterize precisely.

That evaluation is necessary because the common comparison between learned and classical
solvers conflates two contributions. A learned proposal plus refinement is better than a cold
start, which rewards the model for the value of diverse initialization. The control that
isolates that effect is random multi-start at equivalent polish and budget. Applied to our own
system it shrinks our headline claim, and finding where the learned component does pay is this
paper's subject.

MARC is the instrument we measure with: a diffusion denoiser proposes assignments over a
factor graph, an exact CAS supplies the residuals and gradients of a residual-based energy,
deterministic descent on that energy polishes each proposal, and a two-stage symbolic checker
is the sole acceptance criterion. The division of labor between a neural proposer and an exact
symbolic engine is the one AlphaGeometry demonstrated in olympiad geometry
\citep{trinh2024alphageometry}.

Four contributions follow. \textbf{(i)~The structural decision}, as above: the repair ranker
against matched controls, plus the external construction anchor.
\textbf{(ii)~A factorization law.} On the value decision the missing control shrinks the claim
to a single regime, which we then state as a law: a parameter-free factorization in the
measured single-start reachability $q(n)$ reproduces the separable family's best-of-$K$ curve
and identifies the conditions under which a learned proposal can win at all.
\textbf{(iii)~A reusable protocol.} The same discipline applied to our own pilot data uncovers
a trap in how repair studies define their populations: admitting an instance because a
stochastic solver failed on it once lets in instances that succeed under fresh randomness, and
grades every arm on a stream correlated with the criterion that selected them, so any
intervention inherits apparent lift. We state the remedy as a five-step protocol, two-stream
failure selection. \textbf{(iv)~The substrate and its scope}: a pre-registered entrapment result, a partial
cross-family transfer result, and a MATH-benchmark scope measurement we ran on ourselves,
showing autoformalization rather than solving is the binding constraint. Every rate carries
its $N$ and a 95\% Wilson interval or $z$-test; negatives are reported in full.

\section{Related Work}
\label{sec:related}

\paragraph{Multistart global optimization.}
We turn the classical multistart bound into a diagnostic: the measured $\log q(n)$ slope, one
constant reproducing the best-of-8 curve parameter-free, and a two-condition dissection
(collapse \emph{and} separability) that decides in advance when a learned proposal can beat
that bound. The bound itself is textbook, and we use it as such. Best-of-$K$ multistart
succeeds with probability $1-(1-q)^K$ in the single-start reachability $q$, and basin
structure governs the cost of stochastic global search \citep{rinnooykan1987clustering,
rinnooykan1987multilevel}; that annealed noise escapes minima where descent stalls is likewise
standard \citep{welling2011sgld, bras2021langevin, regularized2025langevin}, so our
entrapment result is confirmation on this substrate rather than a finding.

\paragraph{Learned initialization and amortized optimization.}
Our value-proposal negative bears directly on this line, whose standard evaluation is learned
start against cold start: under random multi-start at the same refinement budget, a control it
does not run, our own advantage survives only where solutions factorize per variable and
dimension defeats random search. A model mapping an instance to a solver starting point is a
learned warm start; \citet{amos2023amortized} surveys amortized optimization, learned starts
accelerate AC optimal power flow \citep{baker2019warmstart}, reinforcement learning tunes QP
solvers \citep{ichnowski2021rlqp}, and amortized Langevin inference
\citep{taniguchi2022langevin} and diffusion proposals with bootstrapped refinement
\citep{bootstrapped2025refinement} are the nearest neural instances.

\paragraph{Learning for combinatorial optimization.}
Two branches of neural CO border this work. One is graph diffusion solvers for NP-complete
problems: DIFUSCO \citep{sun2023difusco} and its unsupervised \citep{sanokowski2024diffuco},
trajectory \citep{li2024constraintaware}, and vehicle-routing \citep{anon2026vrpdiffusion}
successors. The other is learning-to-branch, from strong-branching surrogates \citep{khalil2016branch}
to bipartite-GNN branching \citep{gasse2019exact}, surveyed by \citet{bengio2021mlco}. MARC
differs in problem class and in where the verifier sits: continuous algebraic systems, an
exact CAS supplying residuals, gradients, and the sole acceptance gate, and a learned discrete
decision that is a pre-solve repair rather than in-tree branching. Neither branch reports the
budget-matched multistart control this study is built around.

\paragraph{Weak baselines as a known failure mode.}
Where that failure mode is usually diagnosed after the fact, one polish and one checker make
Eq.~\eqref{eq:bestofk} the \emph{exact} bound a learned proposal must clear here, and the
measured $\log q(n)$ slope says in advance which families leave room above it. Neither the
failure mode nor the budget-matching remedy is ours: learned heuristics that do not beat
naive greedy \citep{nath2024maxcutbenchmark}, Tabu Search beating learned local search
\citep{nath2023mightiest}, and the same pattern outside combinatorial search
\citep{rodrigues2026budgetmatched}.

\paragraph{Per-instance algorithm selection.}
The repair ranker is, at bottom, per-instance selection: given an
instance and $K$ discrete options, pick the one under which a downstream solver succeeds.
The problem is fifty years old \citep{rice1976algorithm}, was made standard practice by
SATzilla \citep{xu2008satzilla}, and is surveyed in \citet{kerschke2019survey}; AlphaGeometry
\citep{trinh2024alphageometry} is the precedent for learning
the structural decision itself, and discrete diffusion \citep{austin2021d3pm,
vignac2023digress} supplied the formalism for our withdrawn predecessor policy. Three things here are not in that line: certificate-grade menu semantics (``exactly one
solvable option'' is a CAS theorem, not a budget-relative claim), candidate conditioning (the ranker encodes each
candidate-augmented graph), and controls that price the classical recourse rather than a
single default algorithm.

\paragraph{Distance geometry.}
On the geometric family below we measure under shared machinery what this literature
characterizes structurally: the reachability slope under one fixed polish, and a trained
proposal that ties restart-matched random search there anyway. The structure is not ours.
Chains of points pinned by squared-distance constraints form a distance geometry problem in
which each new point lies on an intersection of circles, the per-point reflection ambiguity is
the discrete branch that branch-and-prune enumerates, and the solution set carries a group of
partial reflections \citep{lavor2012dmdgp, liberti2014distance}; those branches are exactly
the ambiguities our reachability measurement counts.

\section{Method}
\label{sec:method}

A problem instance is a factor graph $G = (V, F)$ over variable nodes $V = \{x_1, \dots,
x_n\}$ and factor nodes $F = \{g_1, \dots, g_m\}$ carrying symbolic expressions: a bare $g$
denotes the equality $g = 0$, relational expressions denote non-strict inequalities. The
residual $r_j(x)$ is $|g_j(x)|$ for equalities and the hinge $\max(0, g_j(x))$ for
inequalities, so $r_j(x) = 0$ exactly when constraint $j$ holds. The energy is
\begin{equation}
E(x) \;=\; \tfrac{1}{2} \sum_{j=1}^{m} r_j(x)^2 .
\end{equation}
Residuals, energy, and $\nabla E$ are computed exactly by a CAS (SymPy) and compiled once per
graph. Acceptance is a two-stage gate: a numeric stage rejects fast if any factor's violation exceeds
a tolerance, and a symbolic stage snaps the candidate to nearby exact rationals and re-checks
every constraint. On the synthetic families the symbolic stage is
authoritative. On the real systems, whose roots are irrational, no exact rational exists to snap to
and the numeric stage stands alone. This gate is the only acceptance criterion in every
experiment and the only reward source in training.

\paragraph{The learned proposal.}
The forward process is standard Gaussian diffusion over variable values with a cosine noise
schedule over $T = 1000$ steps. The denoiser is a standard bipartite message-passing GNN: variable
nodes are encoded from their noisy value and type, factor nodes from their type,
residual, and a sinusoidal timestep embedding; $L$ rounds of message
passing are followed by a per-variable head predicting the noise
$\hat{\varepsilon}$. Sampling runs DDIM reverse steps with energy-gradient guidance,
\begin{equation}
\hat{s} \;=\; s_\theta(x_t, t, G) \;-\; \lambda_t \nabla E(x_t),
\end{equation}
where $s_\theta$ is the score implied by the predicted noise ($s_\theta \propto
-\hat{\varepsilon}$); the exact CAS gradient steers the reverse process toward feasibility.
One architectural detail proved necessary: a linear skip from incident-factor constants
to the variable's output, without which per-round LayerNorm washes out pinned values
(mean absolute root error 5.4 versus 0.9; Appendix A of the supplement).

\paragraph{Propose and polish, and the baseline battery.}
The system solver is a hybrid: the diffusion model proposes $K$ candidate assignments
(best-of-8 throughout); each is polished by deterministic descent on $E$; the first candidate
to pass the checker is the answer. Each baseline isolates one ingredient: deterministic energy descent ($\sigma_k \equiv 0$ in
$x^{(k+1)} = x^{(k)} - \eta \nabla E(x^{(k)}) + \sigma_k \xi_k$) stalls where
$\nabla E = 0$ but $E > 0$; annealed Langevin ($\sigma_k$ decreasing) is the classical
stochastic solver; the mean-prior control tests whether the model does more than reproduce
an average solution. The decisive control is random multi-start: $K$ initializations drawn
uniformly from a family-matched box, polished identically, same budget, no learning.

\section{Experiments}
\label{sec:experiments}

\subsection{Protocol}
\label{sec:entrapment}

All problem families are procedurally generated with held-out test instances. Unless stated
otherwise, solve rates are best-of-8 over $N = 60$ held-out instances per family, intervals
are 95\% Wilson intervals, and comparisons are one-sided two-proportion $z$-tests of
learned $>$ baseline. Before any comparison we
verified that the learned solver converges at all: after five implementation faults were found
and fixed (Appendix A of the supplement), the solve rate on convex linear systems
went from 0 to 1.000 in-distribution and held out. Convex families are saturated by every
method and carry no comparative signal. All results below are non-convex.

A pre-registered check confirmed that stochasticity matters in
this search space. Deterministic energy descent is trapped on
every one of 200 non-convex instances (rate 1.000), and annealed Langevin cuts entrapment to
0.475, a reduction of $0.525 \pm 0.109$ whose 95\% interval excludes zero (5 seeds). This is
textbook behavior, and we present it as a confirmed premise rather than a finding.

\paragraph{Two-stream failure selection.}
One protocol element is stated in reusable form because a later result lives or dies by it.
Repair-style studies build their population from instances a stochastic pipeline fails.
Conditioning on one bad draw admits instances that succeed under fresh randomness and grades
every control on a stream correlated with the failure criterion, so any intervention inherits
apparent lift. The discipline: (i) admit an
instance only if the direct solve fails on two independent streams; (ii) label interventions
on a third, majority-voted when the labels train a model; (iii) grade all arms on a fourth,
common across arms; (iv) field a matched-budget restart control beside the interventions;
(v) before crediting a learnable selection target, screen each candidate at the target budget
on fresh streams and grade the argmax on a held-out one. If that per-instance measurement
cannot beat the matched-budget control, the advantage is portfolio breadth, not selection a
model could amortize. On our geometry population single-stream selection put the construction ceiling at
$0.63$--$0.74$ against $0.09$--$0.27$ for matched restarts, and two-stream selection with a
common grading stream collapsed the margin, most of it the control returning to its true
rate.

\subsection{Diverse starts and polish account for the hybrid's performance}
\label{sec:hybrid}

On four non-convex families where deterministic descent is fully trapped (best-of-8, 60
held-out instances per family), the learned proposal records no win over random multi-start.
It ties on three families (0.550, 0.683 and 0.683, each matching the random arm exactly) and
fails completely on the fourth, CircleLine, at 0.000 where random reaches 0.200.
Cold-start refinement solves nothing and Langevin noise helps only somewhat, while both the
random-init control and the learned hybrid do far better (supplementary Table S1).
Multi-start Levenberg--Marquardt saturates all four, so the informative contrast is learned
against random, not learned against cold-start Langevin.
Without the random control the ablation reads as ``the
learned solver beats classical refinement''; with it, random multi-start plus deterministic
polish beats cold-start Langevin and the proposal adds nothing, because the solutions are
small integers that random restart hits readily and there is nothing to
amortize. What carries this result is the restart-and-polish recipe, not the denoiser. Cross-family transfer is likewise
partial. The cross-trained hybrid reaches 0.683 on two of four held-out families
($p < 10^{-4}$) and solves nothing on the other two
(Appendix A of the supplement).

\subsection{Dimension scaling: where the learned proposal wins}
\label{sec:scaling}

Where random restart collapses geometrically and solutions are per-variable separable, the
learned proposal holds essentially flat. The families above gave random restart an easy
target, so we use bundled non-convex traps whose roots vary per instance over a wide signed
range ($\pm[3,8]$), so neither a fixed prior nor a lucky draw suffices (best-of-8, unified-v2
protocol: one shared polish and one checker for every arm, $N{=}40$ per cell; the full battery
is supplementary Table S3, and Figure~\ref{fig:law} carries the two arms that
matter).
There is a clean crossover by $n = 3$. At $n = 1$ random restart wins, 1.000 against 0.950.
Its solution space is small enough to brute-force, so learning buys nothing. From $n = 3$
random restart collapses (0.075, then 0.000), because a random initialization must land in
all $n$ basins at once. The learned
proposal holds essentially flat where random fails: 0.950 at $n = 2$, 0.975 at $n = 3$, 0.925
at $n = 4$. That is the ceiling the law below takes as its input.
Langevin and the mean-prior sit at or near zero from $n = 3$. This is the amortized-inference
result, a per-instance proposal replacing search whose cost is exponential in dimension,
with one caveat: the learned arm is seed-unstable at intermediate $n$ (single seed
0.250 at $n = 6$; three seeds read $0.983 \pm 0.014$ there but $0.658 \pm 0.484$ at
$n = 4$). The law's larger-sample measurement below reads random lower at low $n$
and places the crossover earlier, so we report the geometric collapse, not a
single $n^*$.

Against the objections that this is one designed family and random multi-start a weak
baseline: rerun on three structurally different separable families with a
Levenberg--Marquardt arm (analytic Jacobian, eight Gaussian multistarts), the crossover
replicates, and LM is not a way out, since it must also hit all $n$ independent basins and
collapses along the same $v^n$ curve.
On two of the three the learned
proposal significantly beats \emph{both} random restart and LM at high dimension (learned
$1.000$ against $0.000$ for both at $n = 6$ on the baseline family). Crossover is
therefore a property of separable structure, not of one construction.
On the third (two spurious wells) the small denoiser did not learn the harder
marginals and collapsed with the classical methods. That sharpens the boundary. Learning
wins only where it can actually amortize per-variable marginals.

\subsection{A factorization law predicts both results}
\label{sec:law}

The crossover above and the coupled negative reported next are two outcomes of one mechanism,
statable as a law and testable without fitting anything. Every method in
this study shares one polish operator and one acceptance checker and differs only in where
the polish starts. Define the single-start reachability
\begin{equation}
q(n) \;=\; \Pr_{x_0 \sim U}\big[\,\text{polish}(x_0)\ \text{is accepted}\,\big],
\end{equation}
a geometric property of the problem family and the polish operator, with no
learning in it. Random multi-start with budget $K$ succeeds iff at least one of $K$ i.i.d.\
starts is accepted, so
\begin{equation}
P_{\text{random}}(n;K) \;=\; 1 - \big(1 - q(n)\big)^{K} .
\label{eq:bestofk}
\end{equation}
The identity is exact for a fixed instance; across a family we use the instance-averaged
version, and we bound the Jensen gap this opens by validating against a self-measured
best-of-8 curve rather than assuming it negligible.
It already implies that a hybrid beating cold-start Langevin has shown nothing. It must beat
Eq.~\eqref{eq:bestofk} at the same $K$.

How $q(n)$ scales with dimension is decided by whether the acceptance basins factorize across
variables. If each factor touches one variable, the polish is coordinate-decoupled and a start is
accepted only when every coordinate independently lands in its root basin, so $q(n) = v^n$
with $v := q(1)$: $\log q$ is linear in $n$ and random search needs $\Theta(v^{-n})$ starts
to hold a fixed success rate. If factors couple, the solution is a joint object and $q(n)$
need not decay geometrically at all. We measured $q(n)$ directly by single-start polish on 600 fresh instances per $n$, the
largest sample in the study and drawn separately from every solve-rate cell, using the same
generators, polish, and checker as those experiments. supplementary Table S4 tabulates the dichotomy.

\begin{figure*}[t]
\centering
\includegraphics[width=\textwidth]{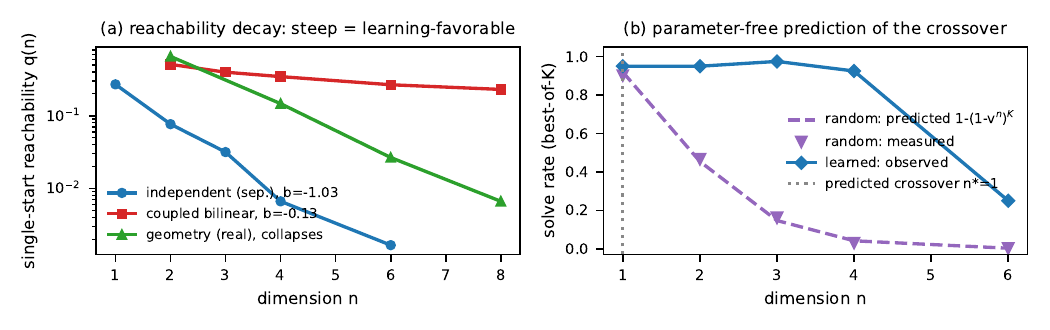}
\caption{The factorization law, measured. Left: $\log q(n)$ against $n$; the separable family
is a line (slope $-1.03$, $R^2 = 0.98$), the coupled bilinear family nearly flat ($-0.13$),
and the geometry family, though syntactically coupled, collapses ($-0.77$). Right: the best-of-8
random-restart curve predicted parameter-free from $v = 0.27$ via Eq.~\eqref{eq:bestofk}
against the measured curve (MAE 0.012), with the learned hybrid overlaid; the dotted line is
the crossover dimension predicted from $v$.}
\label{fig:law}
\end{figure*}

The separable family behaves exactly as the law requires: $\log q(n)$ is linear with slope
$-1.03$ ($R^2 = 0.98$). Substituting the directly measured $v = q(1) = 0.27$ into
Eq.~\eqref{eq:bestofk} reproduces the entire best-of-8 random-restart curve with no free
parameters. It predicts 0.919,
0.454, 0.147, 0.042, 0.003 against measured 0.902, 0.467, 0.162, 0.030, 0.002 at
$n = 1, 2, 3, 4, 6$, a mean absolute error of 0.012
(Figure~\ref{fig:law}; 600 fresh instances, a larger and separately drawn sample than
supplementary Table S3's $N = 40$ random arm). A budget-fair
reading is the expected number of restarts from the \emph{measured} $q(n)$ rather than the
$v^{n}$ extrapolation: $1/q(n)$ runs 3.7 to 600 over the separable dimensions against
2.0--4.3 on the coupled family. Any fixed budget is exhausted on the
former, while on the latter random search never collapses and a learned proposal has
nothing to amortize; an oracle-marginal control, reported with the coupled results below,
confirms this is a property of the family rather than of our model.

\begin{figure}[ht]
\centering
\includegraphics[width=\columnwidth]{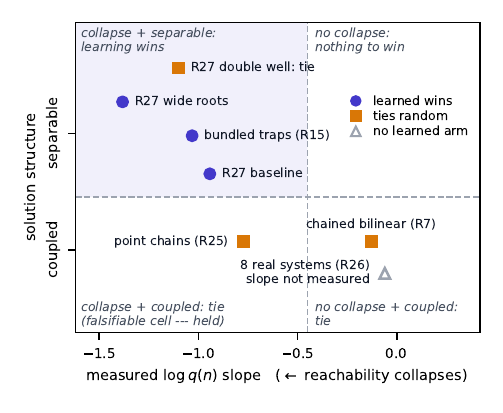}
\caption{The regime map. Each measured family sits at its measured $\log q(n)$ slope
(abscissa) in its solution-structure band (ordinate, categorical), colored by the measured
learned-vs-random outcome; slope provenance is in Appendix C of the supplement. The cells give the law
directly: learning wins only where reachability collapses \emph{and} solutions are
per-variable separable. The chained bilinear family fails collapse and ties; the geometry
point chains collapse but are coupled and tie, which was the falsifiable cell and it held. The
double-well tie inside the win cell is a capacity failure, not a regime failure; the
real-systems suite has no learned arm and no measured slope (nominal abscissa).}
\label{fig:regime}
\end{figure}

Separability is sufficient for geometric collapse but not necessary, so the diagnostic is the
measured slope, not the syntactic label. A real-valued geometric domain makes this concrete:
chains of unknown points ($n = 2k$ coordinates) with squared-distance constraints to fixed
anchors and between consecutive points, a non-convex quartic energy whose solutions the
checker accepts exactly. The family is syntactically coupled, yet its reachability collapses:
slope $-0.77$ ($R^2 = 0.999$), $q(n) = 0.653$, 0.147, 0.027, 0.007 at $n = 2, 4, 6, 8$
(Figure~\ref{fig:law}, left). Each point's two-circle subproblem has a reflection
ambiguity, so a random start rarely places every point in the right basin. Collapse alone
would suggest an amortized proposal wins here. Training a denoiser on this family under the
identical protocol does not bear that out. The learned proposal ties random restart at every
chain length and collapses with it (both 0.625, 0.175, 0.025, 0.000 at $n = 2, 4, 6, 8$,
$N = 40$ per length; zero of four significant wins).
That negative is the informative result, because it separates two conditions the independent
family had conflated. Reachability collapse (condition one, read off the slope) determines
whether random search fails. It does here. Separability (condition two) determines whether the
denoiser has per-variable marginals to amortize. Geometry fails it, because each point's
coordinates are pinned only jointly. Both conditions are necessary, so a learned proposal
beats classical search only where reachability collapses \emph{and} the solution is
per-variable separable; given those, it wins when the denoiser fits the marginals.
Figure~\ref{fig:regime} places every measured family on the two axes of the law.

\subsection{Coupling removes the advantage}
\label{sec:coupled}

The law predicts where the crossover must disappear: when acceptance basins do not factorize,
$q(n)$ stays nearly flat, random restart never collapses, and a learned proposal has nothing
to amortize. The test is the coupled chained-bilinear family, $x_i + x_{i+1} = s_i$ and
$x_i \cdot x_{i+1} = p_i$, whose solution is a joint object and whose measured reachability
slope is $-0.13$ (supplementary Table S4), so what follows is a prediction checked, not
an unexplained negative. Table~\ref{tab:coupled} reports best-of-8 with 60 test instances per
$n$.

\begin{table}[t]
\centering
\caption{Coupled chained-bilinear systems (best-of-8, 60 test per $n$). The learned proposal
records zero significant wins over random restart in five comparisons.}
\label{tab:coupled}
\footnotesize
\setlength{\tabcolsep}{4pt}
\begin{tabular}{ccccl}
\toprule
$n$ & Langevin & random & learned & learned $>$ random? \\
\midrule
2 & 0.167 & 0.483 & 0.233 & no (loses) \\
3 & 0.100 & 0.600 & 0.533 & no \\
4 & 0.050 & 0.517 & 0.517 & no (tie) \\
6 & 0.000 & 0.367 & 0.333 & no \\
8 & 0.000 & 0.467 & 0.483 & no \\
\bottomrule
\end{tabular}
\end{table}

Across all five dimensions the learned proposal records no significant win. That high-dimensional advantage was an independence artifact: it needed
per-variable-separable solutions the model can memorize as marginals, and a random restart
forced to hit all $n$ basins by chance. Coupling removes both, and the proposal adds nothing
over random search plus refinement.

The negative is mechanistic, not a verdict on our denoiser's capacity. Sampling each variable
from its \emph{true} per-variable marginal (pooled from disjoint instances) and polishing identically also
ties random restart at every $n \ge 3$ here (0/4 wins): a proposal that fit the marginals
perfectly would still not beat restarts, because population-marginal information is exhausted
once the solution is a joint object. The claim therefore bounds
product-of-marginals proposals, the class our denoiser belongs to; an instance-conditional
joint sampler is outside what this control tests.

\paragraph{Standard real systems.}
Every family above is procedurally generated, so we also ran the solver battery on eight
recognized test problems encoded once as factor graphs, graded by the numeric stage of the
checker ($\max_j |r_j(x)| < 10^{-6}$) since their roots are irrational
(Table~\ref{tab:real}).
Multi-start Levenberg--Marquardt solves all eight. Where the gradient polish fails, always at
single-start reachability $0.00$, the stronger \emph{classical} polish fixes every case: the
bottleneck is the finisher, not the proposal.
No system falls in the amortization regime, since all eight are low-dimensional and coupled.

\begin{table}[t]
\centering
\caption{The eight standard systems (best-of-8; numeric acceptance at $10^{-6}$).
$q$ is single-start reachability under the gradient polish, 200 draws; $\nabla$ marks the
stationary-point systems. Deterministic descent solves none, and every system the gradient
polish misses sits at $q = 0.00$.}
\label{tab:real}
\footnotesize
\setlength{\tabcolsep}{3pt}
\begin{tabular}{lccccc}
\toprule
System & $n$ & $q$ & Langevin & random & LM \\
\midrule
Circle intersection    & 2 & 1.00 &            & \checkmark & \checkmark \\
Conic--line            & 2 & 1.00 &            & \checkmark & \checkmark \\
GPS trilateration      & 2 & 0.00 &            &            & \checkmark \\
Rosenbrock $\nabla$      & 2 & 0.00 &          &            & \checkmark \\
Himmelblau $\nabla$      & 2 & 0.00 &          &            & \checkmark \\
2R inverse kinematics  & 4 & 0.98 & \checkmark & \checkmark & \checkmark \\
3R inverse kinematics  & 6 & 0.00 &            &            & \checkmark \\
Cyclic-4               & 4 & 0.38 &            & \checkmark & \checkmark \\
\bottomrule
\end{tabular}
\end{table}

\subsection{Relocating the learned component: structural repair beats its controls}
\label{sec:repair}

Moved to the discrete decision, the learned component clears its matched controls decisively.
Coupling closed the route where the network predicts values, but the one decision classical
solvers cannot make is discrete and prior to any solve: \emph{which} structure-changing
augmentation turns an unsolvable graph into a solvable one. A candidate-conditioned
ranker applies each of $K$ proposed augmentations (here $K$ is the menu size, not the restart
budget of Eq.~\eqref{eq:bestofk}), encodes the resulting polynomial graph with operator-aware
node and edge features (degree, linear/square/cross participation, constants), and scores the
augmented graphs listwise; the highest-scoring repair receives one classical solve and the
exact checker remains the acceptance gate. A matched candidate-only control sees the same augmentation recipe but no problem graph;
\texttt{random} chooses uniformly among the same $K$ candidates. The predecessor policy classified menu slots over one fixed graph encoding and fell to chance
on an unseen pattern; candidate conditioning is the change that matters.
An ablation locates the signal: masking operator-identity features and
retraining leaves the ranker intact ($0.978$ $[0.957, 0.989]$ against $0.997$ unmasked,
$N{=}360$; reruns at fixed seed move the third decimal, so we read the interval). The signal lives in constants, magnitudes, and incidence read
\emph{jointly with the candidate}, not in operator flags.

\begin{table}[t]
\centering
\caption{Candidate-conditioned repair ranker against a candidate-only control and random
selection, on three generalization tests (Data Version 8; significance by exact paired McNemar in
the text). Nonlinear menus
carry exact CAS no-real-roots certificates for their distractors; every arm is graded by the
same reference solver that certified the data.}
\label{tab:repair}
\footnotesize
\setlength{\tabcolsep}{3.5pt}
\begin{tabular}{lrccc}
\toprule
Test & $N$ & ranker & cand.-only & random \\
\midrule
Balanced nonlinear      & 360 & \textbf{0.997} & 0.333 & 0.236 \\
Vieta $\to$ unseen rel.\ & 150 & \textbf{0.420} & 0.120 & 0.253 \\
Linear, unseen pattern  & 400 & \textbf{0.380} & 0.195 & 0.287 \\
\bottomrule
\end{tabular}
\end{table}

Table~\ref{tab:repair} reports the three generalization tests under a deliberately strict
protocol: one reference solver certifies the data, grades every arm,
and runs the end-to-end solves; ``exactly one solvable option'' is an exact CAS theorem (a
distractor proven to have no real solution is unsolvable at any budget); gold and distractor
parameters share one support and one prior, so surviving surface-form signal is capped at the
candidate-only ceiling and anything above it must come from the problem graph. On the balanced nonlinear test that ceiling is $0.333$, above the $0.25$ chance floor and
far below the ranker, where exact paired McNemar gives $p = 3.3\times10^{-83}$ against
random (274 ranker-only correct versus 0) and $p = 1.1\times10^{-72}$ against the
candidate-only control. Transfer to an unseen nonlinear relation is partial but bidirectional ($0.420$ against
$0.253$ for vieta-trained on quad\_link, $0.393$ against $0.180$ reversed; $N{=}150$ each). On an unseen linear pattern the ranker reaches $0.380$ against $0.287$ ($N{=}400$); holding
out each of the other two patterns gives $0.407$ and $0.450$ against $\sim$$0.23$ random, so
the effect is not specific to one holdout. A single model tested across all three at once reads $0.339$ against $0.249$
($N{=}1{,}200$, $p = 7.8\times10^{-7}$). The linear signal is reported at its post-audit size: closing each discovered shortcut
lowered it ($0.565 \to 0.445 \to 0.380$ across data versions) while pushing the
candidate-only control to chance, so what remains is problem-graph reading, modest where
every operator is linear and decisive on certificate-grade nonlinear menus. Both headline rows were retrained across three optimization seeds (11, 29, 47) with
independent per-seed evaluation draws; nonlinear comes through optimization-robust
($0.982 \pm 0.006$).
The linear edge is not seed-stable ($0.317 \pm 0.069$, one seed at $0.227$ below the random
arm's $0.248 \pm 0.002$), consistent with reading the linear rows as mechanism evidence
rather than a deployable margin.

Learning is also cheap here where value diffusion was not: one forward pass per candidate
rather than a reverse-diffusion rollout. End to end, the nonlinear ranker's single call solves
$0.933$, exactly the oracle and enumeration ceiling, where blind enumeration averages $2.62$
calls ($N{=}60$); linear $K{=}4$ solves $0.300$ against $0.227$ random with a $1.000$ ceiling
($N{=}300$). A cheap-probe control bounds the claim from the other side: a short-budget call on every
candidate solves at most $0.881$ of nonlinear menus at $4.7$ calls, where
the ranker's single call solves $0.939$ on that
larger $N{=}360$ population (the $0.933$ above is the matched $N{=}60$ end-to-end run). Learning beats probing on accuracy
and cost at once. Provably rootless distractors are unsolvable at any budget, and short
probes miss the gold. On linear menus the probe saturates and enumeration is already perfect at $2.5$ calls.
Two negatives bound the menu-size claim: the $K{=}4$ checkpoint transfers to larger menus
without retraining, but its accuracy advantage is gone by $K{=}16$, and training directly at
$K{=}16$ sits at chance (Appendix E of the supplement). ``Exactly one solvable option'' is exact for all linear menus (rank) and $99\%$ of nonlinear
test menus (CAS real-root nonexistence); the remainder carry a disclosed budget-relative
probe certificate.

\paragraph{The structural decision on systems we did not design.}
Those menus are ours. We therefore hardened four of the named real-system classes into
parameterized variants and asked whether \emph{derived} constructions, functions of the
givens alone such as law-of-cosines lifts and signed Cayley--Menger cross-product pins,
repair the failures the classical reference cannot (supplementary Table S5 reports
all four classes). Two produce a two-stream failure population: far-side GPS trilateration ($0.848 \pm 0.020$
over three seed bases, $N{=}509$ failures) and a ghost-root conic--line intersection
($0.263 \pm 0.015$, $N{=}158$). On 3R inverse kinematics and far circles the classical solver
never fails, which we report as a negative rather than averaging in as a zero. On the two that do produce failures, one construction chosen on a disjoint half of each pool
repairs \emph{every} held-out instance ($1.000 \pm 0.000$ across seeds; pooled Wilson
$[0.99, 1.00]$ and $[0.98, 1.00]$) against $0.433 \pm 0.049$ and $0.114 \pm 0.011$ for a
restart control matched to the \emph{full enumeration budget} (McNemar $p < 10^{-13}$; the
control, not the treatment, carries the sampling variance). The same
construction wins in all six folds, which rules out its being the luckiest of $V$, and
saturation is
mechanical rather than fortunate: the selected pins delete the mirror basin and the ghost
root by construction. No learned model runs here. What transfers to systems we did not
design is the value of the structural decision, not the ranker.

\paragraph{Where the construction result does not extend.}
This result holds where the failure is a systematic attractor that one construction deletes
outright, and not where failures are stochastic. On the pruned point chains, the
discrete-branch setting of distance geometry, the population definition decides the answer. Under two-stream selection ($N{=}367$ failures, three optimization seeds) the enumeration
ceiling is $0.692$ at $72.7$ restarts per instance, which plain restart scaling matches at
$+32$ ($0.725$). Majority-vote labels made the target learnable, and the ranker separates from random
($0.246 \pm 0.016$ versus $0.185 \pm 0.019$; Holm $p{=}1.3{\times}10^{-4}$) where
single-stream labels never did, but that is no edge over the controls: it ties the best fixed
construction ($0.259$) and loses to matched-budget restarts ($0.270$). The signal is real but reflects the population prior, not the problem graph. The probe initially appeared promising: one restart per candidate solves $0.698$, matching
the ceiling. But it spends the whole menu's budget and its accepts are diffuse, so it is a
portfolio sweep, not a choice of construction. A per-instance screen held to the reference
budget scores $0.199$ $[0.161, 0.243]$, below the restart control, so ${\sim}50$ restarts of
measurement select worse than the prior. The learned arms landing on the prior is therefore expected. Scoring that same screen on
the streams that chose it reads $0.762$, which prices the selection-on-noise effect. A
construction pays when it removes a named failure mode outright; where none does,
constructions help only as a portfolio and ``which construction'' carries no signal a
selector can exploit.

\section{Limitations}
\label{sec:limitations}

Our positive result has a stated scope. The external anchor uses \emph{derived} constructions
and a held-out choice among them, not the learned ranker; its populations are hardened
variants chosen to expose a failure mode, and two of the four produce no failure population. We measure the ranker itself on menu-based repair over synthetic factor
graphs.

On the value side, the learned proposal's advantage requires per-variable-separable
solutions and a regime where random restart collapses; on coupled systems it never
significantly beats random restart at any dimension tested. It gains nothing at $n = 1$ (random restart at ceiling), and single cells are seed-noisy, so
the useful window is the crossover region, not high dimension per se. Those families are also
block-decomposable: a solver that read off separability would scale linearly there too, so
the crossover is demonstrated against joint-start search, not block decomposition. Eq.~\eqref{eq:bestofk} predicts the random-restart curve, not the learned ceiling, which
is measured.

Several failures are specific. CircleLine is never solved by the learned proposal, transfer
succeeds on two of four held-out families, and adding CircleLine to the training mix
collapsed an otherwise recoverable transfer (0.70 to 0.00): a bad training family actively
hurts. The predecessor structure policy never
separated from random selection and its numbers remain withdrawn (evaluation seeds
overlapped checkpoint-selection validation seeds), as do the earlier v6/v7 ranker numbers
(a pin-prior leak and probe-artifact certificates, found and fixed). Only the nonlinear headline ($0.982 \pm 0.006$) is optimization-robust.
Finally, scope against real problem text is quantified: on a 48-problem sample of
MATH-500 \citep{hendrycks2021math} the template formalizer covers 0 of 48 (20\%
constraint-shaped, 31\% CAS computation, 48\% reasoning or proof out of
scope). The binding constraint is autoformalization, not the solver.

\section{Conclusion}
\label{sec:conclusion}

The decision classical solvers make only by enumeration is which structure to add, not which
values to try, and that is where the learned component pays: the repair ranker matches the
enumeration ceiling at a fraction of the calls where ``exactly one solvable option'' is a
theorem, and on hardened real systems a derived construction repairs failures the full
restart budget misses. On the value decision the protocol returns a characterization
instead. The learned proposal pays only where search is separable and dimension defeats
random restart, and the advantage disappears under coupling. The generalizable lesson is
about evaluation: populations defined by stochastic failure are artifacts of the draw unless
the protocol stabilizes them, at one extra solve per instance.

\section*{Use of AI Assistants}
A large language model assisted with the readability of this paper. All experiments,
analysis, and claims are the authors' own; no result, table, or figure was model-generated.

\bibliography{refs}

\onecolumn
\appendix
\setcounter{secnumdepth}{1}
\input{marc_aaai_appendix}
\end{document}

%% file: marc_aaai_appendix.tex

\section{Per-Family Hybrid Battery and Cross-Family Transfer}
\label{app:hybrid}

The five implementation faults fixed before any comparison (the entrapment experiment): the denoiser never
received $x_t$; constants were absent from the graph tensors; $t$ did not condition variable
nodes; guidance magnitudes exploded; and a shape-handling fault in the one-variable case.
After the fixes, the solve rate on convex linear systems went from 0 to 1.000 in-distribution
and held out. A sixth, architectural fix is referenced from the Method section: per-round
LayerNorm washes out the magnitude of constraint constants, so a variable whose value is
pinned by its own constraint cannot be read off the deep features; a direct linear skip from
a variable's incident-factor constants to its output restores the path and cut the mean
absolute root error from 5.4 to 0.9.

Table~\ref{tab:hybrid} is the full mechanism ablation behind the hybrid-recipe study: four non-convex
families where deterministic descent is fully trapped (best-of-8, 60 held-out instances per
family; 95\% Wilson CIs shown on the random-init arm). The random-init control uses identical polish and budget with no
learning.

\begin{table}[ht]
\centering
\caption{Mechanism ablation on trapped non-convex families (best-of-8, 60 held-out per
family; 95\% Wilson CIs on the random-init arm; columns are cold-start refinement, refinement with Langevin noise, random-init plus polish, multi-start Levenberg--Marquardt, and the learned hybrid). Values are those of the committed
\texttt{results/p\_hard/hard\_eval.json}. The multi-start Levenberg--Marquardt column is
reported for completeness: the strong classical solver saturates all four families, which is
why the learned-versus-random comparison, not the learned-versus-Langevin one, is the
informative contrast.}
\label{tab:hybrid}
\footnotesize
\setlength{\tabcolsep}{4pt}
\begin{tabular}{lccccccl}
\toprule
Family & cold & +Langevin & random-init & LM & learned & learned $>$ random? \\
\midrule
BilinearSystem  & 0.000 & 0.300 & 0.550 [0.42, 0.67] & 1.000 & 0.550 & tie ($p{=}0.50$) \\
BilinearProduct & 0.000 & 0.100 & 0.683 [0.56, 0.79] & 1.000 & 0.683 & tie ($p{=}0.50$) \\
QuadraticSystem & 0.000 & 0.300 & 0.683 [0.56, 0.79] & 1.000 & 0.683 & tie ($p{=}0.50$) \\
CircleLine      & 0.000 & 0.033 & 0.200 [0.12, 0.32] & 1.000 & 0.000 & no (random wins) \\
\bottomrule
\end{tabular}
\end{table}

\label{app:transfer}
Cross-family transfer is partial. We train on three non-convex families and test the hybrid
on the held-out fourth (best-of-8, 60 test instances per family, 200 epochs);
Table~\ref{tab:transfer} reports the outcome. On two of four held-out families the
cross-trained model reaches 0.683, significantly above the refine+Langevin baseline
($p < 10^{-4}$): the model solves a family it never trained on. On the other two it solves
nothing. The BilinearSystem failure is the informative one, because that family is solvable
in-distribution (0.550 in Table~\ref{tab:hybrid}). A smoke run trained on only the two
similar families \{Product, Quadratic\} recovered it at 0.70, whereas adding the pathological
CircleLine family to the training mix collapsed it to 0.00. Dissimilar training families can
degrade transfer. We read this as transfer across related non-convex structure, not
as a universal solver.

\begin{table}[ht]
\centering
\caption{Leave-one-family-out transfer (best-of-8, 60 test per family, 200 epochs). $p$ from
a $z$-test of learned (cross) $>$ refine+Langevin.}
\label{tab:transfer}
\small
\begin{tabular}{lllll}
\toprule
Held-out family & Trained on & refine+Langevin & learned (cross) & $p$ \\
\midrule
BilinearProduct & Sys, Quad, Circle  & 0.100 [0.05, 0.20] & 0.683 [0.56, 0.79] & $<10^{-4}$ \\
QuadraticSystem & Sys, Prod, Circle  & 0.300 [0.20, 0.43] & 0.683 [0.56, 0.79] & $<10^{-4}$ \\
BilinearSystem  & Prod, Quad, Circle & 0.300 [0.20, 0.43] & 0.000 & fails \\
CircleLine      & Sys, Prod, Quad    & 0.033 [0.01, 0.11] & 0.000 & fails \\
\bottomrule
\end{tabular}
\end{table}

\section{Dimension Scaling: Full Battery}
\label{app:scaling}

Table~\ref{tab:scaling} is the full five-arm battery behind the dimension-scaling study; the main text and
Figure 1 of the main paper carry the two arms that decide the claim (random restart and learned).

\begin{table}[ht]
\centering
\caption{Dimension scaling on bundled traps with per-instance-varying roots in $\pm[3,8]$
(best-of-8, unified-v2 protocol: one shared polish and one checker for every arm, $N{=}40$
per cell; 95\% Wilson half-widths are at most $0.15$). Bold marks the best method per
row. A three-seed rerun of the same protocol ($N{=}120$ per cell,
scaling\_3seed.json) reads $0.983 \pm 0.014$ at $n{=}6$ and $0.658 \pm 0.484$ at
$n{=}4$: single cells are seed-noisy, the crossover is not.}
\label{tab:scaling}
\begin{tabular}{cccccc}
\toprule
$n$ & deterministic & Langevin & mean-prior & random restart & learned \\
\midrule
1 & 0.000 & 0.775 & 0.000 & \textbf{1.000} & 0.950 \\
2 & 0.000 & 0.225 & 0.000 & 0.725 & \textbf{0.950} \\
3 & 0.000 & 0.000 & 0.000 & 0.075 & \textbf{0.975} \\
4 & 0.000 & 0.000 & 0.000 & 0.000 & \textbf{0.925} \\
6 & 0.000 & 0.000 & 0.000 & 0.000 & \textbf{0.250} \\
\bottomrule
\end{tabular}
\end{table}

\section{The Factorization Dichotomy: Regime Map and Table}
\label{app:law}

The regime map (Figure 2 of the main paper) places every measured family on the
two axes of the law; Table~\ref{tab:law} tabulates the dichotomy plotted in
Figure 1 of the main paper (600 fresh instances per $n$, Wilson CIs, $K = 8$). Expected restarts
$1/q(n)$ are listed over the tested dimensions in increasing order.

Slope provenance for the regime map: the separable and coupled abscissae are the fits of
Table~\ref{tab:law}, the geometry abscissa is the fit of Figure 1 of the main paper, and the
three crossover-replication families are placed at slopes inverted from their best-of-8
Levenberg--Marquardt arm through the best-of-$K$ identity of the main paper. The real-systems suite is plotted
at a nominal abscissa: it has no learned arm and no separately measured $q(n)$.

\begin{table}[ht]
\centering
\caption{The factorization dichotomy, measured.}
\label{tab:law}
\small
\begin{tabular}{lll}
\toprule
Quantity & Independent (separable) & Coupled \\
\midrule
$\log q(n)$ slope & $-1.03$ & $-0.13$ \\
fit $R^2$ & 0.98 & 0.96 \\
measured $v = q(1)$ & 0.27 & --- \\
$E[\text{starts}] = 1/q(n)$ & $3.7 \to 13 \to 32 \to 150 \to 600$ & $2.0 \to 2.5 \to 2.9 \to 3.8 \to 4.3$ \\
$P_{\text{random}}$ MAE, parameter-free & 0.012 & law N/A (not separable) \\
\bottomrule
\end{tabular}
\end{table}

\section{External Anchor: Per-Class Construction Repair}
\label{app:anchor}

Table~\ref{tab:anchor} is the per-class breakdown behind the external-anchor paragraph.
Both restart controls rerun the same classical reference solver on the unrepaired system:
\texttt{restart} at the reference solver's own budget, \texttt{restart}$\times|V|$ at the
full enumeration budget over the construction vocabulary. The two classes that never fail
are reported as negatives rather than averaged in as zeros; the own-budget column shows how
far short a plain retry falls even before the budget is matched.

\begin{table}[ht]
\centering
\caption{Derived-construction repair on hardened variants of named real systems: three
independent seed bases, $N{=}200$ instances per class per seed, mean $\pm$ population SD.
``Failures'' is the two-stream failure rate of the classical reference and $N$ the pooled
failure population it defines; the construction is chosen on a disjoint half of that pool
and graded on the held-out half.}
\label{tab:anchor}
\small
\begin{tabular}{lrrccc}
\toprule
Class & Failures & $N$ & construction (held out) & restart & restart$\times|V|$ \\
\midrule
Trilateration, far side & $0.848 \pm 0.020$ & 509 & $\mathbf{1.000 \pm 0.000}$
  & $0.079 \pm 0.017$ & $0.433 \pm 0.049$ \\
Conic--line, ghost root & $0.263 \pm 0.015$ & 158 & $\mathbf{1.000 \pm 0.000}$
  & $0.026 \pm 0.010$ & $0.114 \pm 0.011$ \\
3R inverse kinematics & $0.000$ & 0 & --- & --- & --- \\
Circles, far apart & $0.000$ & 0 & --- & --- & --- \\
\bottomrule
\end{tabular}
\end{table}

\section{Repair: Menu-Size Scaling and Cost Accounting}
\label{app:kscaling}

\begin{figure}[t]
\centering
\includegraphics[width=0.85\textwidth]{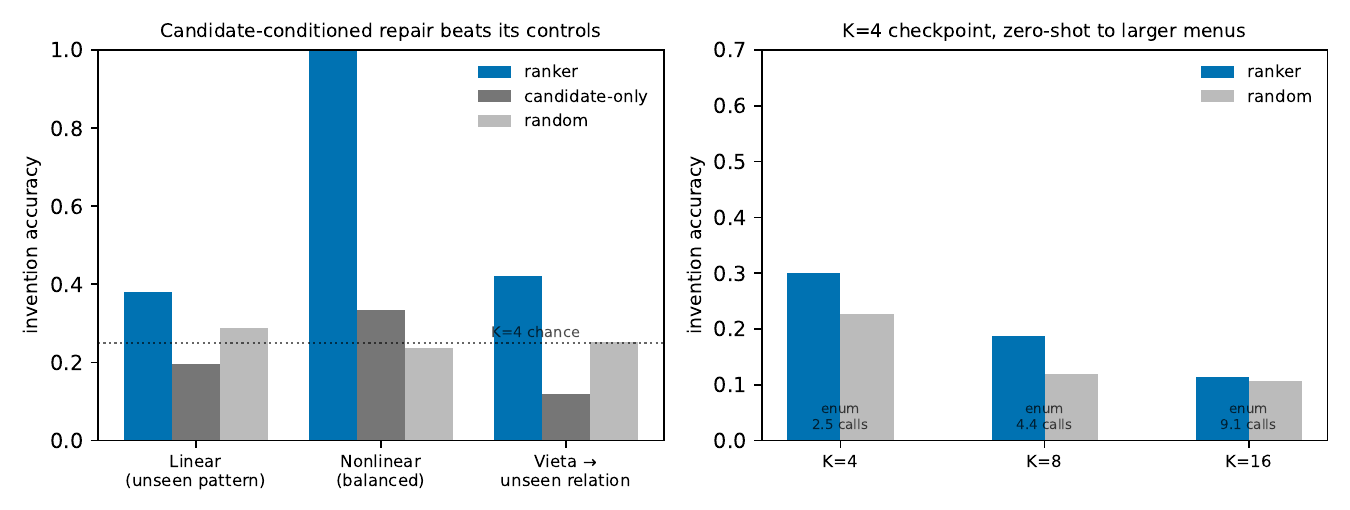}
\caption{Left: Table 3 of the main paper drawn, the ranker against its
candidate-only and random controls, with $K{=}4$ chance dotted. Right: menu-size scaling.
The $K{=}4$ checkpoint evaluated zero-shot at larger menus retains an accuracy edge at
$K{=}8$ that closes by $K{=}16$, while the blind enumeration it displaces grows from $2.5$
to $9.1$ solver calls per instance.}
\label{fig:repair}
\end{figure}

The $K{=}4$ checkpoint transfers to larger linear menus without retraining, but the accuracy
advantage shrinks with $K$ and is gone by $K{=}16$ ($0.300/0.187/0.113$ against random
$0.227/0.120/0.107$ at $K{=}4/8/16$, $N{=}300/150/150$); what survives at large $K$ is only
the cost gap, as the enumeration it displaces grows to $9.05$ solver calls per instance
(Figure~\ref{fig:repair}). Directly training at $K{=}16$ performs at chance, so the
transferred checkpoint is selected and both negatives are kept in the record.

\section{MATH-500 Scope Measurement}
\label{app:math}

The main text states that the template formalizer covers none of a 48-problem MATH-500
sample \citep{hendrycks2021math}. The breakdown: roughly $20\%$ of the sample is
constraint-shaped and in principle in scope, $31\%$ is CAS computation rather than
constraint solving, and $48\%$ is reasoning or proof that the factor-graph encoding does
not express. Coverage is $0/48$ because the constraint-shaped fraction still requires
autoformalization from problem text, which the template formalizer does not perform. The
binding constraint on applying this substrate to natural problem statements is therefore
autoformalization, not the solver.